\documentclass{article}
\usepackage{ijcai19}

\usepackage{float}
\usepackage{times}
\usepackage{soul}
\usepackage{url}
\usepackage[hidelinks]{hyperref}
\usepackage[utf8]{inputenc}
\usepackage[small]{caption}
\usepackage{graphicx}
\usepackage{amsmath}
\usepackage{booktabs}
\usepackage{algorithm}
\usepackage{algorithmic}
\usepackage{latexsym}
\usepackage{amssymb}
\usepackage{multirow}
\usepackage{setspace}
\usepackage{subcaption}

\newcommand{\vect}[1]{\boldsymbol{#1}} 

\title{Learning Structured Embeddings of Knowledge Graphs \\
with Adversarial Learning Framework}

\author{
Jiehang Zeng\and
Lu Liu\And
Xiaoqing Zheng\footnote{Contact Author}\\
\affiliations
School of Computer Science, Fudan University, Shanghai, China\\
\emails
\{jhzeng18, l$\_$liu15, zhengxq\}@fudan.edu.cn
}

\begin{document}

\maketitle

\begin{abstract}
Many large-scale knowledge graphs are now available and ready to provide semantically structured information that is regarded as an important resource for question answering and decision support tasks. However, they are built on rigid symbolic frameworks which makes them hard to be used in other intelligent systems. We present a learning method using generative adversarial architecture designed to embed the entities and relations of the knowledge graphs into a continuous vector space. A generative network (GN) takes two elements of a (subject, predicate, object) triple as input and generates the vector representation of the missing element. A discriminative network (DN) scores a triple to distinguish a positive triple from those generated by GN. The training goal for GN is to deceive DN to make wrong classification. When arriving at a convergence, GN recovers the training data and can be used for knowledge graph completion, while DN is trained to be a good triple classifier. Unlike few previous studies based on generative adversarial architectures, our GN is able to generate unseen instances while they just use GN to better choose negative samples (already existed) for DN. Experiments demonstrate our method can improve classical relational learning models (e.g. TransE) with a significant margin on both the link prediction and triple classification tasks.
\end{abstract}

\section{Introduction}
Knowledge graphs store semantic information in the form of entities and relationships that is easily machine-processable --- a property that is considered as an important ingredient to build more intelligent systems by taking advantage of such semantically structured representations. Thanks to long-term collaborative efforts, many knowledge graphs such as WordNet \cite{acm-miller:95}, YAGO \cite{www-suchanek:07}, DBpedia \cite{iswc-auer:07}, and Freebase \cite{sigmod-bollacker:08}, which contain a huge amount of data, are now readily available and have been successfully used for coreference resolution \cite{acl-ng:02}, question expansion \cite{vldb-graupmann:05}, and questing answering \cite{ai-ferrucci:10}. However, their underlying rigid symbolic representations, while being very interpretable and efficient for their original purposes, make them hard to be integrated, especially into deep learning systems that focus on learning distributed representations of data \cite{aaai-bordes:11}.

A promising method is to embed the entities and relations from a knowledge graph into a continuous low-dimensional vector space. Once those embeddings are well learned from the existing facts, the relationships between entities can be derived from interactions of their embeddings via an appropriate operator for each relation. Many possible ways have been proposed to model these interactions and to derive the existence of a relationship from them \cite{aaai-bordes:11,icml-nickel:11,www-nickel:12,nips-jenatton:12,aaai-bordes:13,nips-socher:13,aaai-wang:14,aaai-lin:15,acl-guo:15,ijcai-xie:16,zhang2018asynchronous}. Knowledge graph completion (or link prediction) is considered as an outstanding merit for these relational learning models since knowledge graphs are often missing many facts, and some of the edges they contain might be incorrect. Encoding entities in distributed embeddings also leads to great improvement in efficiency for the link prediction because such predictions can be made without exploring the original big graph.

The relational learning models for knowledge graphs usually predict the existence of a (subject, predicate, object) triple via a score function which represents the model's confidence that a triple is true. These models are normally trained by maximizing the plausibility of observed triples. However, training on all-positive samples is tricky, because the model easily over generalize \cite{arxiv-nickel:15}. The problem is that knowledge graphs usually only contain positive triples. A widely-used method to generate negative samples is to ``perturb'' positive triples by replacing the subjects or objects of true triples with entities selected at random. Unfortunately, good ``plausible'' negative examples are still hard to come and usually not sufficient to train useful models: while it is relatively easy to predict that a person is born in a city, it is difficult to predict which city in particular. A better approach (based on perturbation) to generate more informative negative examples is to replace the subjects or objects of observed triples with those semantically close to the replaced one.

We propose a relational learning method using generative adversarial architecture \cite{nips-goodfellow:14} in which the negative examples are partly generated by a generative network (GN), and a discriminative network (DN) is trained to distinguish ground truths from the generated triples and randomly sampled false ones. GN and DN compete in a two-player minimax game: the discriminator tried to differentiate the positive triples from the others, and the generator tries to fool the discriminator. Competition in this game drives both networks to improve their performance until the generated examples are indistinguishable from the true triples. When arriving at a convergence, GN recovers the training data and can be used for knowledge graph completion, while DN is trained to be a good triple classifier. Unlike previous work \cite{cai2017kbgan,wang2018incorporating} using generative adversarial architectures, our GN is capable of  unseen ``plausible'' triples whereas they just use GN to grade and select negative samples (already existed) for DN. Experiments showed that our method can significantly improve the performance of classical relational learning models (e.g. TransE) on both the link prediction and triple classification tasks.

The remainder of this paper is structured as follows. Section $2$ presents a brief overview of related work. In Section $3$, our neural network architecture and training algorithm are represented. Section $4$ reports experimental results. The conclusion will be given in Section $5$.

\section{Related Work}

Relational learning methods for modelling knowledge graphs fall into two categories: graph feature-based  \cite{ml-lao:10,emnlp-lao:11,dmkd-rettinger:12,ijcai-wang:16,acl-toutanova:16} and latent feature-based \cite{aaai-bordes:11,icml-nickel:11,nips-socher:13,aaai-lin:15,acl-guo:15} models. The intuition behind the former is that the edges can be recovered by the features extracted from the observable properties of the graph, and those models look at the direct correlation of patterns observed in a graph. The former can be further divided into three classes: those that predict links using path ranking algorithms, those that infer new links using the rules extracted from graphs, and those that link the entities using the derived similarity between them. The latter try to find the correlation between nodes or edges in a graph through latent variables. We here focus on the latter that is more related to this study.

What all latent feature-based models have in common is that they use latent features of entities to explain observable triples, and those features are not directly observed in the data (that is why we call them ``latent''). A (subject, predicate, object)  triple is usually represented by $(h, r, t)$ while the latent features of triples is represented by three vectors $(\vect{h},\vect{r},\vect{t}), \vect{h},\vect{r},\vect{t} \in \mathbb{R}^k$, where $k$ is the dimensionality of latent feature vector representations. The key intuition behind such models is that the relationships between entities can be inferred from the interactions of their latent features. We briefly review several typical ways to model these interactions for predicting the new facts below.

The structured embedding (SE) model \cite{aaai-bordes:11} derives the probability of relationships from the distances between latent feature representations of entities, and model the score of a triple as $f^{\text{SE}}(h, r, t) = ||\vect{v}_{k}^{lhs} \vect{h} - \vect{v}_{k}^{rhs}\vect{t}||$, where the matrices $\vect{v}_{k}^{lhs}$ and $\vect{ v}_{k}^{rhs}$ transform the feature vectors of entities to model relationships specifically for the relation $r_k$. To reduce the number of parameters in SE model, Bordes et al \shortcite{aaai-bordes:13} proposed TransE model that translates the latent feature representations using a relation-specific distance instead of linear transformation. The score of a triple is then defined as $f^{\text{TransE}}(h, r, t) =  ||\vect{h} + \vect{r} - \vect{t}||_2^2$. The main shortcoming of this model is that the latent features of two entities do not interact with each other, because they are independently mapped to a common space. TransH \cite{aaai-wang:14} projects the entity vectors  $\vect{h},\vect{t}$  onto the relation hyperplanes to alleviate  many-to-many problem. Immediately after TransH, TransR \cite{aaai-lin:15} and TransD \cite{Ji2015KnowledgeGE} view entities and relations as two independent space, different mapping technique from entity space to relation space are proposed in these models. Our architecture use those models as basic building blocks, and we show that their performance can be significantly improved using our training method.

RESCAl \cite{icml-nickel:11} is a bilinear model that explains triples by capturing the pairwise interactions between two entity feature vectors using multiplicative terms. The score of a triple is modeled as $f^{\text{RESCAL}}(h, r, t) =  \vect{h}^{\top} \vect{W}_k \vect{t}$, where $\vect{W}_k \in \mathbb{R}^{d \times d}$ is a weight matrix that specifies how the latent features interact for the relation $r_k$. Such bilinear model has been augmented with diagonal weight matrices \cite{iclr-yang:15}, and complex-valued embeddings \cite{trouillon2016complex}. RESCAL can be seen as a special case of tensor factorization methods \cite{siam-kolda:09}, and similar methods have been explored for predicting triples \cite{acm-drumond:12} and modeling highly multi-relational data \cite{nips-jenatton:12}. Socher et al \shortcite{nips-socher:13} stated that the bilinear models can only capture linear interactions and might be unable to fit more complex relations, and they use a neural tensor to directly relate the two entity feature vectors across multiple dimensions. Convolutional neural networks were also tried to capture the similar relations \cite{dettmers2018conve,nguyen2017novel}. Even though neural tensors or convolutional networks have much more expressive power that is useful for modelling large knowledge bases, they have more parameters than SE and RESCAl models. Dong et al \shortcite{sigkdd-dong:14} and Yang el al \shortcite{iclr-yang:15} reported that such models tend to overfit, at least on the relatively small datasets.

Our method is more related to \cite{cai2017kbgan} and \cite{wang2018incorporating} where an adversarial learning framework is applied. Their generators are trained to provide better negative samples for the discriminators than randomly selectors. Their generated negative samples in fact exist in the training dataset, while our generator can produce unseen  ``plausible'' examples. Beside, they train the generator by a reinforcement learning because discrete sampling steps prevents gradients from back-propagating. In our method, the generator was desired to take two elements of a triple as input and make up the missing entity in form of its vector, which makes the whole process easily trainable and fully differentiable.

 
\section{Adversarial Learning-Based Framework}

We here describe an adversarial learning-based framework to model the relation between two entities in a graph through latent variables, in which the entities are embedded into a continuous vector space, and the relations between them can be recovered from their distributed representations (or embeddings). The generative network (GN) is trained to deceive the  discriminative network (DN) by gradually improving its ability in generating ``just-like-truth'' triples, while DN is taught to differentiate truth triples from the generated ones as well as randomly selected negative samples. Competition in this game drives both networks to improve their performance until the generated triples are indistinguishable from the genuine ones. In our framework, any relational learning model (e.g. TransE) can play the role of GN or DN in such two-player game, and we try to explore these two possibilities. In this section, we first formally introduce the architecture of the proposed framework, and then describe two implementations of this framework: one taking a translation-based model as the generator, and another using it as the discriminator.

\begin{figure} 
  \includegraphics[width=8.5cm]{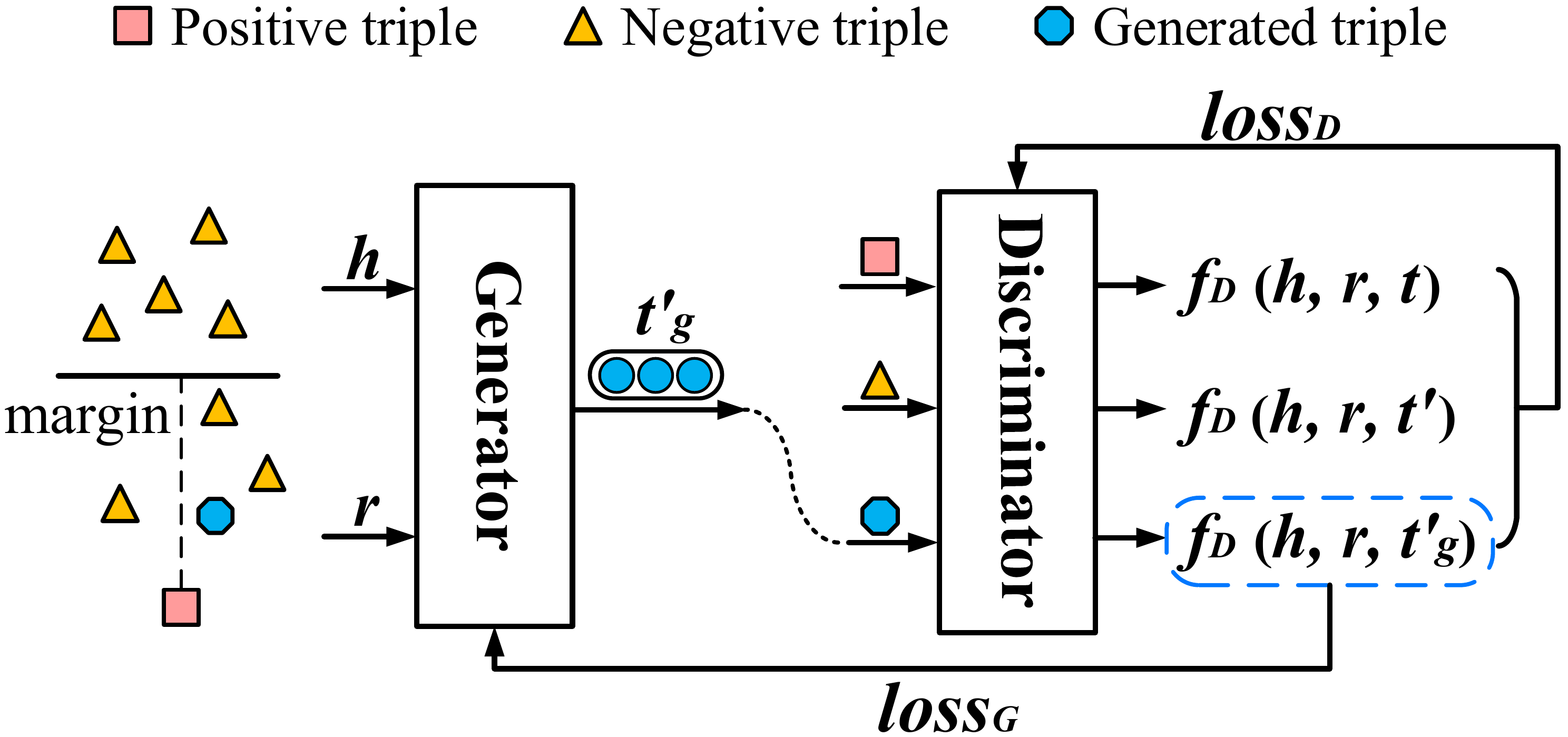}  
  \caption{An overview of our framework. The generator (GN) takes the vector representations of a head $h$ and a relation $r$ as input, and produce the feature representation of a ``plausible'' $t'_{g}$. The discriminator (DN) receives the generated triple $(h,r,t'_{g})$, the ground truth $(h,r,t)$ and randomly selected negative sample $(h,r,t')$, and scores those triples via a function $f_D(\cdot)$ to distinguish the truth triple from the others. The training goal of GN is to deceive DN to make wrong classification. When arriving at a convergence, GN can be used for knowledge graph completion, while DN is taken as a triple classifier. When a tail is given to GN, we convert it to a head by using the reverse of the input relation.} 
  \label{figure1}
\end{figure}

\subsection{Architecture}
A knowledge graph (KG) consists of a set of triples $(h, r, t) \in S$, where $h, t \in \mathcal{E}$ are entities and $r \in \mathcal{R}$ is a relation. We call $h$ the head, and $t$ the tail of a triple $(h, r, t)$ that represent $h$ has the relation $r$ with $t$. To embed the entities and relations of KG into a continuous vector space, an adversarial learning-based framework is designed to learn such embeddings so that the triples stored in KG can be recovered from those embeddings. The proposed framework is illustrated in Figure \ref{figure1}, which has two main components: a generator (GN) and a discriminator (DN). GN takes a head-relation pair $(h, r)$ as input, and attempt to generate the vector representation of a tail $t'_{g}$ that should be indistinguishable from the truth tail $t$. DN is trained to distinguish the ground truth triples from others by a score function $f_D(\cdot)$, and the score of a triple represents its confidence that the triple is true. 



The training objective of DN is to differentiate the ground truth triples $\left( h, r, t \right)$ from the triple $\left( h, r, t'_{g} \right)$ generated by GN. However, in the early stages of training, GN is incapable of generating good negative samples for DN because GN is not well trained yet. Thus, another negative sample $\left( h, r, t'\right)$ is added to train DN, and $(h, r, t')$ is a false triple constructed by replacing the correct tail with $t'$ randomly sampled from the entities in KG. Inspired by Wasserstein GAN \cite{arjovsky2017wasserstein}, the loss function for DN's part can be formalized as below, where $f_{D}\left (h ,r , t\right)$ is a scoring function. 
\begin{equation} \small \label{loss_DN}
\begin{aligned}
L_{D}\   = \sum_{( h, r, t)\in S } \{2f_{D} (h ,r , t) - f_{D}(h, r,   t'_{g}) - f_{D}(h, r, t')\}
\end{aligned}
\end{equation}
\noindent In our definition, the lower the score $f_{D}\left (h ,r , t\right)$ is, the more likely the triple $(h,r,t)$ is true. As training progresses, DN learns to becomes a better triple classifier. The score of positive sample, $f_{D} (h ,r , t)$, is doubled in order to counteract the effect of two types of negative samples. 

GN is trained to generate the embedding of $t'_{g}$ for a given $(h, r)$ pair, and to make DN taking $(h, r, t'_{g})$ as a truth triple. Thus, the training loss of GN is defined as follows.
\begin{equation} \small \label{loss_GN}
\begin{aligned}
L_{G}\  = \sum_{( h, r, t ) \in S } \{f_{D} (h, & r, t'_{g})\}
\end{aligned}
\end{equation}
When arriving at a convergence, GN learns to generate the ``plausible'' triples that are indistinguishable from the genuine ones, which ability can be used for knowledge completion.

GN and DN are trained jointly in a two-player minimax game, and they use the same embedding matrices $\vect{E} \in \mathbb{R}^{n\times k }$ (for entities) and $\vect{R} \in \mathbb{R}^{m\times k }$ (for relations), where $n$ is the number of entities, $m$ the number of relations, and $k$ is the dimensionality of embeddings.

As mentioned above, any relational learning model can be served as the role of GN or DN. However, if overly complex models are used at the both sides, they may suffer from a very large search space, which makes them difficult to be trained, especially for the adversarial learning situation. We reduce the search space by requiring at least one of GN or DN to adopt simple, but robust translation-based model, such as TransE \cite{aaai-bordes:13}, TransH \cite{aaai-wang:14}, and TransD \cite{Ji2015KnowledgeGE}. We explore several variants for the proposed framework and mainly build two types of implementations: one using translation-based models as DN while another taking them as GN. 


\subsection{Translation-based discriminator} \label{sec:DN}
In this setting, we choose to use multilayer perceptron (MLP) or convolutional neural network (CNN) as the generator, and one of translation-based models (TransE, H, and D) is taken as the discriminator, so we have six different combinations. 

When playing as a generator, CNN takes the concatenation of vector representations of a head $h$ and a relation $r$ as input, A convolution with multiple filters is used to yield another feature vector by taking the dot product of filter vectors with the input vector. After the input vector is convoluted with the filter matrix, a non-linear function is applied, following a classical linear transformation. A MLP is a class of traditional feed-forward neural network, consisting of multiple linear layers, interleaved with some non-linearity function. 


When taking a translation-based model as the discriminator, we need to add a marginal term $\gamma$ into Eq. (\ref{loss_DN}) like TransE \cite{aaai-bordes:13}, and the loss is rewritten as:
\begin{equation} \small \label{loss_DN_Trans}
\begin{aligned}
L_D = \sum_{\left( h, r, t \right) \in S} & [ 2f_{D}\left( h, r, t \right) - f_{D}\left(h, r, t'_{g} \right) \\ & - f_{D}\left(h, r, t' \right) +\gamma  ]_{+}  
\end{aligned}
\end{equation}
\noindent The discriminator works as a maximum-margin classifier so that the distances between positive and negative samples are maximized with the chosen hyperplane.


\subsection{Translation-based generator} \label{sec:GN}
A translation-based model is used as the generator in this configuration. Taking TransH \cite{aaai-wang:14} as example, the generated triple in its vector representation takes the form of $\left( \vect{h}_{\bot}, \vect{r}, \vect{h}_{\bot} + \vect{r} \right)$, where $\vect{h}$ and $\vect{r}$ denote the embeddings of a head $h$ and a relation $r$ respectively, and $\vect{h}_{\bot} = \vect{h} - \vect{w}^{\top}_r\vect{h}\vect{w}_r$. The weights in $\vect{w}_r$ are used to project the vector $\vect{h}$ onto the hyperplane, defined for the relation $r$. 

When taking a translation-based model as a generator, the loss function for GN's part is defined as:
\begin{equation} \small \label{loss_Trans}
\begin{aligned}
L_{G}\ =  \sum_{( h, r, t)\in S } &  \{[f (h ,r , t) - f(h,  r,   t') + \gamma  ]_{+} \\
& + f_{D} (h, r, t'_{g} )\}
\end{aligned}
\end{equation}
\noindent This loss has two parts: one is defined to maximally separate the positive triples from the negative ones in the hyperspace, and the other is used to better deceive the discriminator by making use of its feedback. As discussed before, various neural networks, including MLP and CNN, are tried as discriminative networks to test several variants for the framework.

\section{Experiments}
We evaluate our adversarial learning-based framework on two standard tasks for learning structured embedding of KG: link prediction and triple classification. For the link prediction, we report the results produced by generators, while for the triple classification, the performance of discriminators is reported. 

\subsection{Datasets}
We used WN18RR \cite{dettmers2018conve} and FB15k-237 \cite{fb15k} as datasets for the link prediction, and WN11 \cite{nips-socher:13} and FB13 \cite{nips-socher:13} for the triple classification. WN18RR and Fb15k-237 are built to remove the reversible relations existed in WN18 and FB15k, which are much easier to be predicted. WN18RR is a subset of WN18 after removing such relations, and Fb15k-237 is a subset of FB15k by removing the redundancy. For each relation $(h, r, t)$, we add its reverse relation $(t, r\_rev, h)$ into the dataset so that our GN always takes the a head-relation pair as input when it is fed with a $(r, t)$. For example, the triple $(LosAngeles, LocatedIn, California )$ is expanded to $(California, LocatedIn\_rev, LosAngeles)$. However, we make sure that if a relation is in the training set, its reverse must not occur in the test.

A set of negative triples is required to evaluate the triple classifier. We choose to use the datasets released by Socher et al \shortcite{nips-socher:13}, where one negative sample are added for each positive triple. The size of datasets is listed in Table $1$. 

\begin{table}[!htbp]  
\centering
\begin{tabular}{l|rcrrr}
\hline
\hline
Dataset & Entity & Rel & Train & Valid & Test \\
\hline
WN18RR & $40,943$ & $11$ & $86,835$ & $3,034$ & $3,134$\\
FB15k-237 & $14,541$ & $234$ & $272,115$ & $17,535$ & $20,466$\\
WN11 & $38,696$ & $11$ & $112,581$ & $2,609$ & $10,544$\\
FB13 & $75,043$ & $13$ & $316,232$ & $5,908$ & $23,733$ \\
\hline
\hline
\end{tabular}
\caption{The size of datasets.}
\end{table} \label{tb:dataset}

\begin{table*}[!htbp]
\centering
\begin{tabular}{cc|ccc|ccc}
\hline
\hline
 \multicolumn{2}{c|}{\multirow{2}{*}{Model}} & \multicolumn{3}{c|}{\textbf{WN18RR}} & \multicolumn{3}{c}{\textbf{FB15k-237}} \\
 ~ & ~ & MR & MRR & Hit@10 & MR & MRR & Hit@10 \\
\hline
\multicolumn{2}{l|}{DistMult$^{\S} $ \  \cite{iclr-yang:15}} & $5110$ & $0.430$ & $49.0$ & $254$ & $0.241$ & $41.9$ \\
\multicolumn{2}{l|}{ComplEx$^{\S}$ \   \cite{trouillon2016complex}} & $5261$ & $\textbf{0.440}$  & $51.0$ & $339$ & $0.247$ & $42.8$ \\
\multicolumn{2}{l|}{KBGAN$_{1}$ \  \cite{cai2017kbgan}} & - & $0.214 $ & $47.2$ & - & $0.278$ & $45.8$ \\
\multicolumn{2}{l|}{KBGAN$_{2}$ \  \cite{cai2017kbgan}} & - & $0.215$ & $46.9$ & - & $0.277$ & $45.8$ \\
\multicolumn{2}{l|}{ConvE \   \cite{dettmers2018conve}} & $4187$ & $0.430$ & $\textbf{52.0}$ & $244$ & $\textbf{0.325}$ & $\textbf{50.1}$ \\
\hline
 \multicolumn{2}{l|}{TransE$^{\dagger}$ \  \cite{aaai-bordes:13} } & $3924$ & $0.178$ & $45.1$ & $197$ & $0.256$ & $41.9$ \\
 \multicolumn{2}{l|}{ GN (MLP) + DN (TransE)} & $1789$ & $0.206$ & $49.1$ & $218$ & $0.244$ & $41.8$   \\
 \multicolumn{2}{l|}{ GN (CNN) + DN (TransE)} & $2970$ & $0.207$ & $49.1$ & $204$ & $0.248$ & $41.4$ \\
 \multicolumn{2}{l|}{ GN (TransE) + DN (MLP)} & $2350$ & $0.220$ & $49.9$ & $188$ & $0.283$ & $47.4$ \\
 \multicolumn{2}{l|}{ GN (TransE) + DN (CNN)} & $2317$ & $0.223$ & $50.1$ & $193$ & $0.276$ & $47.5$ \\
\hline
\multicolumn{2}{l|}{TransH$^{\dagger}$ \ \cite{aaai-wang:14}} & $4113$ & $0.186$ & $45.1$ & $202$ & $0.231$ & $40.1$ \\
 \multicolumn{2}{l|}{ GN (MLP) + DN (TransH)} & $\textbf{1555}$ & $0.205$ & $48.8$ & $220$ & $0.242$ & $41.1$ \\
 \multicolumn{2}{l|}{ GN (CNN) + DN (TransH)} & $2050$ & $0.193$ & $46.3$ & $221$ & $0.223$ & $38.7$ \\
 \multicolumn{2}{l|}{ GN (TransH) + DN (MLP)} & $1970$ & $0.235$ & $51.2$ & $193$ & $0.282$ & $47.7$ \\
 \multicolumn{2}{l|}{GN (TransH) + DN (CNN)} & $1899$ & $0.234$ & $\textbf{52.0}$ & $191$ & $0.280$ & $48.0$ \\
\hline
 \multicolumn{2}{l|}{TransD$^{\dagger}$ \ \cite{Ji2015KnowledgeGE} } & $3555$ & $0.190$ & $46.4$ & $188$ & $0.245$ & $42.9$  \\
 \multicolumn{2}{l|}{ GN (MLP) + DN (TransD)} & $2585$ & $0.194$ & $46.0$ & $270$ & $0.222$ & $36.9$ \\
 \multicolumn{2}{l|}{ GN (CNN) + DN (TransD)} & $3963$ & $0.184$ & $43.9$ & $258$ & $0.229$ & $37.3$  \\
 \multicolumn{2}{l|}{GN (TransD) + DN (MLP)} & $2515$ & $0.216$ & $49.7$ & $178$ & $0.287$ & $46.7$ \\
 \multicolumn{2}{l|}{GN (TransD) + DN (CNN)} & $2685$ & $0.220$ & $50.0$ & $\textbf{173}$ & $0.292$ & $47.5$ \\
\hline
\hline
\end{tabular}
\caption{The experimental results for the link prediction task. MR, MRR and Hit@$10$ denote the mean rank, mean reciprocal rank and Hits with tenth (in \%) respectively. The results indicated with $\S$ are excerpted from \protect\cite{dettmers2018conve} and those indicated with $\dagger$ are from \protect\cite{zhang2018nscaching}. The results of KBGAN$_{1}$ and KBGAN$_{2}$ were obtained by the two best combinations (TRANSD + DISTMULT) and (TRANSD + COMPLEX) reported in \protect\cite{cai2017kbgan}. Our results are listed below each baseline system used as a building component, where GN indicates which model is used as the generator and DN indicates which as the discriminator in our adversarial learning-based framework, like ``GN (MLP) + DN (TransE)''.}
\end{table*}

\subsection{The Choice of Hyperparameters}
We use grid search technique to determine the values of hyper-parameters from few choices: $k \in \left\{50, 100, 200 \right\}$ for the dimensionality of embeddings, $\gamma \in \left\{0.5, 1, 2 \right\}$ the margin, $\eta \in \left\{0.001, 0.0005, 0.0001 \right\}$ the learning rate, $B \in \left\{1000, 5000, 10000 \right\}$ the batch size, $\lambda \in \left\{0, 0.00001 \right\}$ the weight decay, $n_{critic} \in \left\{ 1, 3, 5 \right\}$ the number of critic iterations for the used Wasserstein GAN, and $c \in \left\{ 0.01, 0.05, 0.1 \right\}$ the clipping threshold. We test the MLP with different layers $l \in \left\{ 1, 2, 3 \right\}$ and hidden sizes $h \in \left\{ 100, 200, 2000, 4000 \right\}$. As to the convolutional network, we explore several number of filters $\tau \in \left\{ 100, 200, 500 \right\}$. 

We tuned the hyper-parameters on the validation dataset, and use RMSProp optimizer to update the parameters of networks, which is also recommended in Wasserstein GAN \cite{arjovsky2017wasserstein}. The L2-norm is used to initialize the embeddings of entities and relations.



\subsection{Link Prediction}

Link prediction aims to predict the missing entity $h$ or $t$ for a positive triple $\left( h, r, t \right)$. We evaluate the performance following the ``filtered'' setting \cite{aaai-bordes:13}: ranking the test triples against all corrupted triples except the test triplet of interest not appearing in the training, validation, or test sets. We employ three widely-used evaluation metrics: Mean Rank (MR), Mean Reciprocal Rank (MRR) and Hits with tenth (Hits@$10$). The lower MR, the higher MRR, and the higher Hits@$10$, the better.

On the validation set of WN18RR, the highest scores in Hits@$10$ is achieved with $k = 100$, $\gamma = 1$, $\eta = 0.001$, $B = 5000$, $\lambda = 0.00001$, $n_{critic} = 1$, $c = 0.01$ and $\tau = 100$. We achieved the highest performance on FB15k-237 with the similar values of hyperparameters except for setting $\gamma$ to $0.5$. 

The results of the link prediction are shown in Table 2 by comparing to the baseline or four state-of-art systems. Our results are all achieved by the generators for this task. We listed several different implementations for the framework, where GN indicates which model is used as the generator and DN indicates which as the discriminator in our adversarial learning-based framework. For example of ``GN (MLP) + DN (TransE)'', this implementation uses a MLP as the generator and takes a model based on TransE as the discriminator.

From these numbers, a handful of trends are readily apparent. First, the adversarial learning framework improves all the performance of
baseline systems in most cases. The proposed method boosts the baseline systems by $11.38\%$ and $12.64\%$ in average of Hit@$10$ on WN18RR and FB15k-237 respectively. The largest increment is achieved by ``GN (TransH) + DN (CNN)'' with $15.3\%$ improvement in Hit@10 and $25.9\%$ improvement in MRR. Another striking result of these experiments is that ``GN (TransH) + DN (CNN)'' achieved state-of-the-art on the WN18RR, and comparable result on FB15k-237 in the metric of Hit@$10$, using relatively quite simple models as its components.


Although ``(TransH) + DN (CNN)'' does not outperform ConvE on FB15k-237 in Hit@$10$ and MRR, it performs competitively. The main goal of this study is to investigate how well the proposed framework can further improve the baseline systems (such as TransE, D and H) comparing with the other adversarial learning-based method. The best result on FB15k-237  achieved by ConvE \cite{dettmers2018conve} with relatively high computational cost. For each training triple, ConvE needs to compute the dot product of a tail vector with all the entity embeddings in KG that might be not scale well to large knowledge graphs. Furthermore, we tried only a few different network configurations, and there are many ways (such as unsupervised pre-training, and carefully designed network architecture) that we could improve it further. ``GN (TransH) + DN (CNN)'' performed better than KBGAN \cite{cai2017kbgan} with a significant margin because the reinforcement learning is not required to train the networks, and our entire training process is fully differentiable even though the similar adversarial learning-based framework is applied.

\subsection{Triple Classification}
Triple classification aims to judge how well a classifier to distinguish the a ground truth triple from the others. Given a triple, a specific threshold $\delta_{r}$ is required for the classifier to decide whether the triple is true. Specifically, if the score of $f_D(h, r, t)$ is less than the given threshold $\delta_{r}$, the triple $(h, r, t)$ is classified as a fact, otherwise a false. $\delta_{r}$ is chosen by maximizing classification accuracy on the validation set. We reported the performance of triple classification using the outputs of the discriminators. The experimental results are shown in Table $3$.
On the validation set of WN11, the highest accuracy is obtained by using $k = 100$, $\gamma = 1$, $\eta = 0.001$, $B = 5000$, $\lambda = 0.00001$, $n_{critic} = 1$, $c = 0.01$ and $\tau = 100$. We achieved the highest performance on FB13 with the similar values of hyperparameters except for setting $\eta $ to $0.0005$.

\begin{table}[!htbp] \small
\centering
\begin{tabular}{l|c|c}
\hline
\hline
Model & WN11 & FB13 \\
\hline
TransE $\left(unif.\right)$ \cite{aaai-bordes:13} & $75.9$ & $70.9$ \\
TransE $\left(bern.\right)$ \cite{aaai-bordes:13} & $75.9$ & $81.5$ \\
TransH $\left(unif.\right)$ \cite{aaai-wang:14} & $77.7$ & $76.5$ \\
TransH $\left(bern.\right)$ \cite{aaai-wang:14} & $78.8$ & $83.3$ \\
KG2E$^{\dagger}$ \cite{he2015learning} & $85.4$ & $\textbf{85.3}$ \\
\hline
GN (CNN) + DN (TransE) & $83.0$ & $79.2$ \\
GN (TransH) + DN (CNN) & $\textbf{85.5}$ & $83.6$ \\
\hline
\hline
\end{tabular}
\caption{The results for the triple classification, reported in accuracy ($\%$). The results of TransE and TransH are extracted from \protect\cite{aaai-wang:14}, in which ``unif'' denotes the way that the negative samples are produced by replacing a head or a tail with a randomly selected entity, while ``bern'' denotes the similar sampling method but the entity is selected according to their frequencies \protect\cite{aaai-wang:14}. The results indicated with $\dagger$ are excerpted from \protect\cite{xiao2016transg}.}
\end{table}

The results of TransE and TransH are excerpted from \cite{aaai-wang:14}, ``unif" denotes the way that the negative samples are produced by replacing a head or a tail with a randomly selected entity, while ``bern" denotes the similar sampling method but the replacing entity is picked according to their frequencies \cite{aaai-wang:14}. As shown in Table $3$, ``GN(CNN) + DN(TransE)'' improves TransE by $7.1\%$ in accuracy on WN11, and ``GN(TransH) + DN(CNN)'' boosts TransH by $6.7\%$ on FB13. ``GN (TransH) + DN (CNN)'' achieved the highest accuracy $85.5\%$ on WN11 dataset and comparable accuracy $83.6\%$ on FB13 datasets, comparing with the listed competitors.


\section{Conclusion}
We proposed a novel generative adversarial-based framework to learn structured embeddings of knowledge graphs, in which the generator is trained to recover the training data while the discriminator is trained to be a good triple classifier. Experimental results demonstrate that our method can improve classical relational learning models (e.g. TransE, D, and H) with a significant margin on both the link prediction and triple classification tasks. Unlike few previous studies based on generative adversarial architectures, our generative network is able to generate unseen instances while they use it as just a negative sample selector for the discriminative ones. The ability in directly generating the feature vector representations of unseen ``plausible'' entities make the framework promising for practical integration with other intelligent systems, especially for deep learning-based systems that focus on learning distributed representations. Such distinguishing feature is analogous to the open-world assumption of description logics with respect to the modeling languages developed in the study of databases.

\small {
\bibliographystyle{named}
\bibliography{ijcai19}
}

\end{document}